# Learning from demonstrations: An intuitive VR environment for imitation learning of construction robots

Kangkang Duan; Zhengbo Zou *

Department of civil engineering, The University of British Columbia, Vancouver, BC, Canada.

*Corresponding author, email: zhengbo@civil.ubc.ca

**Abstract**

Construction robots are challenging the traditional paradigm of labor intensive and repetitive construction tasks. Present concerns regarding construction robots are focused on their abilities in performing complex tasks consisting of several subtasks and their adaptability to work in unstructured and dynamic construction environments. Imitation learning (IL) has shown advantages in training a robot to imitate expert actions in complex tasks and the policy thereafter generated by reinforcement learning (RL) is more adaptive in comparison with pre-programmed robots. In this paper, we proposed a framework composed of two modules for imitation learning of construction robots. The first module provides an intuitive expert demonstration collection Virtual Reality (VR) platform where a robot will automatically follow the position, rotation, and actions of the expert's hand in real-time, instead of requiring an expert to control the robot via controllers. The second module provides a template for imitation learning using observations and actions recorded in the first module. In the second module, Behavior Cloning (BC) is utilized for pre-training, Generative Adversarial Imitation Learning (GAIL) and Proximal Policy Optimization (PPO) are combined to achieve a trade-off between the strength of imitation vs. exploration. Results show that imitation learning, especially when combined with PPO, could significantly accelerate training in limited training steps and improve policy performance.

1. **Introduction**

On-site construction work involves lots of repetitive, dangerous, and heavy-load tasks (such as material transportation and structural assembly), which make the construction industry both labor- and mechanically-intensive [1]. This has also created a need for automatic construction techniques such as Building Information Modeling (BIM) and robotics [2-3]. Recent years have seen remarkable potential for construction robots, which can be defined as programmable, active mechanisms capable of a variety of construction tasks. Specifically, construction robots have shown improved efficiency in earthmoving [4-5], structural assembly [6-9], waste recycling [10-11], and wall mortar spraying [12].

An automated construction robot is expected to perform tasks without human intervention and work in dynamic environments [1], which brings challenges to pre-programmed robots [13]. Pre-programmed construction robots [7,14-16], known for their reliability and predictability [9], are the early form of construction robots. They are provided with the precise trajectory before performing a task (e.g., block assembly [7], tile installation [16]), and they follow pre-programmed instructions, incapable of changing the trajectory during execution. However, this inability to adapt to the constantly changing and complex construction environments hinders the applicability, and therefore, the wide adoption of pre-programmed construction robots onsite.

RL provides a promising solution to this problem [17-18]. RL aims to learn a policy, often represented by a deep neural network, that maps the states (e.g., the positions and joint angles of robots) of construction robots to their actions [19]. During training, the robot updates its policy by exploring the environment and seeking high rewards, so an optimal policy is expected to allow the robot to maximize the expected reward. Related studies [19-20] have shown that, for each state $s$, there does exist at least one optimal policy that can maximizes the expected future reward that a robot can obtain at state $s$ in the stationary Markov decision processes. Since RL tries to train a policy network by exploring the action and state space to determine the optimal action, construction robots based on RL are more adaptive to environmental changes [21]. Recent studies have reported success in adopting RL control methods such as Q-learning and PPO in construction tasks such as pick and place [22], and window installation [9].

If the action and state spaces are continuous and high dimensional, however, it is computationally expensive to adequately explore the action and state spaces to arrive at an effective policy. Without any prior knowledge, the robot will aimlessly executing random actions until it finds a valid trajectory in the high dimensional space [23]. Another problem for RL-based construction robot revolves around the reward function. Many construction sequences (e.g., concrete pouring and structural assembly) include several subtasks. For some tasks, it is reasonable that the agent will receive a reward only if the task is completed [9]. For example, an assembly robot needs to finish two subtasks. First, it needs to pick up an object, and then it needs to place the object at the target position. It is natural to reward the robot only if it had successfully picked or placed the object. This kind of rewarding mechanism (i.e., sparse reward) is difficult for traditional RL algorithms to learn since getting positive feedback is rare in a high-dimensional space [23-25].

A feasible approach to address these two issues is IL [20,26-27]. IL is developed based on conventional RL theorem [20,28-29]. In IL, the robot directly learns from expert demonstrations, thus reducing the efforts wasted on exploration. Some IL methods train a policy network with state observations as the input and expert actions as the ground truth output, which is referred as BC [26-27,30], while others try to recover the expert policy by finding a reward function under which the expert policy is the optimal policy, which is referred as inverse reinforcement learning (IRL) [20,28,31-33]. Both BC and IRL need expert demonstrations as inputs. These demonstrations record the observation and action of the expert, so it is necessary to provide an immersive or realistic construction environment to collect the expert's instinctive actions.

IL has been proven effective in complex task learning in various environments [34]. However, current implementations of expert demonstration collection are still limited in terms of demonstration diversity, safety of demonstrators, cost, and reproducibility. Given that VR has shown significant advantages in the development and simulation of realistic construction environments, in this study, we utilized VR to facilitate demonstration collection used for IL of construction robots (Fig. 1). The main contributions of our work are as follows:

- We proposed an intuitive demonstration collection platform where a construction robot will precisely track the actions of an expert in VR and save the action and observation data for imitation training.
- We utilized the demonstrations with IL to solve the challenges of sparse reward, and also combined the PPO method so that the robot can discover more efficient policies.
- We investigated the factors (e.g., the size of demonstration datasets) that could affect the performance of IL controlled construction robots through tests in a robot simulation platform.

The structure of this paper is: first, we discussed related work in the use of RL and IL in construction robots and applications of VR in construction robots; second, we introduce the implemented approaches for the first module, i.e., the intuitive demonstration collection module; third, the implemented approaches of the second module (IL-based control module), as well as the experimental task, were discussed; finally, we discussed the effectiveness of the entire framework and the influencing factors of IL.

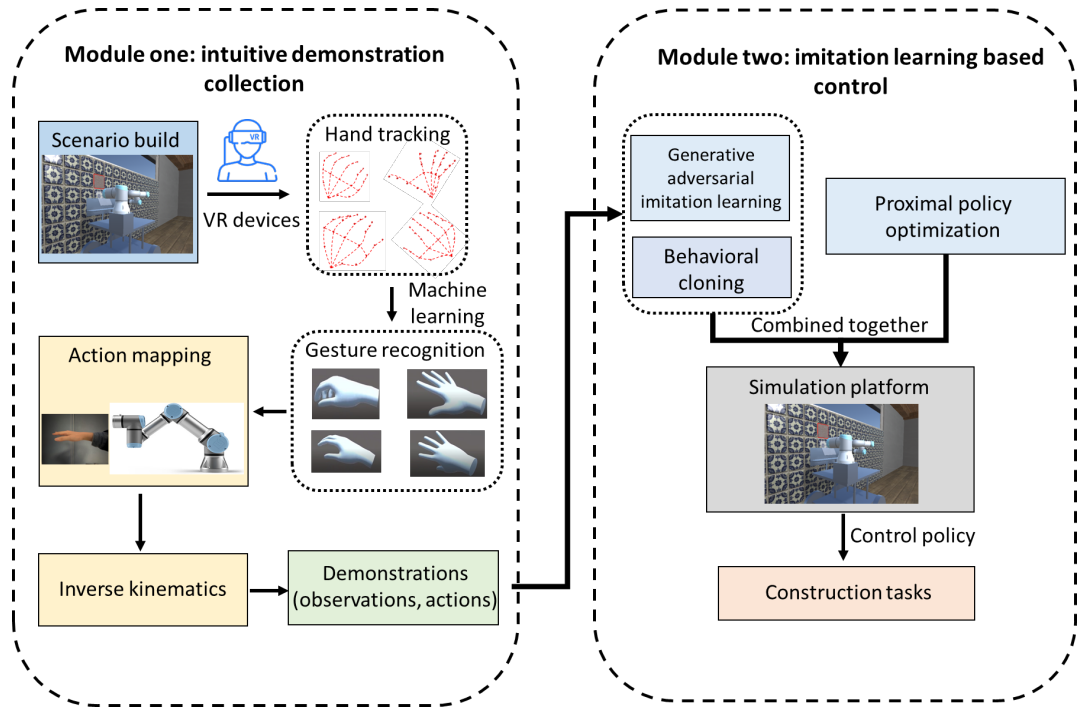

Fig. 1 The workflow of the framework: Module one provides an intuitive demonstration collection platform where the robot will follow the action, position, and rotation of the expert's hand. Module two combines IL and PPO, and uses the demonstrations (the observations and actions) collected from module one to train a construction robot.

2. **Background**

**2.1 Reinforcement learning in construction robots**

Driven by sufficient exploration and exploitation in the action and state spaces, RL aims to propose a flexible and adaptive policy to maximize the reward function [19]. Recent studies have reported its success in many fields such as manufacturing [35-36] and self-driving [37]. However, in construction robots, the application of RL [8-9,21-22,38-42] is still an emerging topic. Recent research works have investigated the application of RL in common construction tasks such as structural assembly [8-9,38-39], earth-moving [40], and hoist control [21]. These studies have pointed out that RL can provide a more efficient policy (e.g., reducing the waiting time and lifting time for a hoist control system [21]) in comparison with manual operations. Belousov et al. [39] adopted deep deterministic policy gradient [43] (DDPG) to train a 6 degree of freedom robot arm to carry a building module to the target position. Apolinarska et al. [38] used DDPG [43] to control a 6 degree of freedom robot arm to complete the insertion task in structural assembly.

These trajectory planning tasks are appropriate to provide a continuous reward feedback while the robot is executing the task (e.g., the robot gets more rewards as it approaches the object), which makes the problem solvable for many RL methods such as DDPG and PPO. When the pick and place actions are involved, the task become more challenging since even though the robot is close to the object it cannot complete the task until it picks up the object, i.e., the reward only occurs after the object is picked. Sparse reward problems are challenging because it is difficult for the robot to continuously get feedback to improve its policy [44]. In addition, even for widely-studied trajectory planning tasks, the design of the reward function still can be complex and needs careful consideration [39]. In this study, we define the task as a sparse reward problem to simplify the design of the reward function and adopt IL to reduce the random exploration of RL algorithms.

**2.2 Imitation learning in construction robots**

The goal of IL is to extract a policy network from expert demonstrations so that a robot will behave like an expert. BC and IRL are two main branches of IL. BC uses supervised learning to train a policy network, where the states of expert demonstrations are used as input and expert actions are the outputs. Since BC only focuses on state action pairs in the demonstrations, a large demonstration dataset is essential. Furthermore, BC generalizes poorly in circumstances where the current state has not been observed by the supervised learning algorithm in demonstration [45], which causes issues in dynamic environments such as construction sites.

IRL has greater generalization ability in comparison to BC as it aims to learn a reward function from demonstrations and then generate the optimal control policy using regular RL from that reward function. Andrew et al. [20] suggested that the reward function could provide a much more parsimonious description of behaviors of the agent. In addition, the presumption of RL is that the reward function is the most succinct and robust definition of the task [20]. Generally, the reward function is chosen from a reward function family (e.g., Gaussian processes [46], neural networks [47]), so careful tuning is required and the reward function family (since we want to approximate the true reward function via a reward function lying in the family) should be expressive enough.

Even though studies [34,46-47] have shown the advantages of IL in solving long-horizon tasks (a task where the reward is sparse and delayed over a long period of time [44]), the application of IL in construction robots is still limited [13,17]. Liang et al. [8] used trust region policy optimization (TRPO) as the training algorithm to generate the policy for ceiling tile installation tasks, and the goal of TRPO is to minimize the difference between the expert observation and the robot observation. Lei et al. [9] added expert trajectories to the demonstration trajectory buffer of the policy gradient algorithm so that the construction robot can learn a suboptimal policy at the early stage of training, which reduced the time needed to discover valuable actions.

Both BC and IRL require demonstrations. However, there are several problems in terms of the collection of demonstrations. First, expert trajectories may not be the perfect trajectories for tasks, which might limit the efficiency of construction robots if only learned through BC or IRL. The second problem is concerned with the delay between the action and the observation due to the use of simulation platforms, which creates a disconnect between the true state observation and the expert demonstrations [26]. These problems might further impact the performance of IL. Therefore, in this paper, we made a trade-off between exploration and exploitation by combining IL with PPO. In addition, given the weak generalization ability of BC, BC was only used in the pretraining stage (i.e., BC can only affect the training in the first 100,000 steps).

**2.3 VR in construction robots**

VR has been widely applied to create construction scenarios for operation and training of construction robots due to concerns about safety risks, cost, and disturbance of work on-site [48]. Studies [8-9,48-55] have demonstrated the success of VR in realistic construction scenario development. For example, Pooya et al. [49] utilized VR to improve workers' ability in working with construction robots. This kind of VR-based training [51-53] is proven to be effective and economical in fostering trust in robots and reducing the mental workload in human-robot interaction. Zhou et al. [54] proposed a robot teleoperation interface where an immersive construction scenario model was built from point cloud data and provided for robotic teleoperation. In robotic teleoperation, VR provides an immersive environment for human operators, so that they can control the construction robot to execute on-site tasks and get feedback in real-time in the VR environment. The construction robot can either directly execute the human actions [48,55] or get high-level task planning from experts [50].

In IL, VR also provides a platform for demonstration collection. Studies [8-9] have shown the feasibility of using VR to collect expert demonstrations. However, previous collection methods require the expert to manipulate the robot using VR controllers instead of directly using their body movements, so the actions collected are not the instinctive operations of humans which might include delay and bring difficulties to operations. In this paper, we introduced a more intuitive approach towards demonstration collection where the hand movements of demonstrators were automatically mapped to the actions of construction robots.

**3. Intuitive demonstration collection module**

This module provides a platform for intuitive demonstration collection. An expert is asked to complete construction tasks in a virtual environment. In this environment, the actions of a human body are mapped to the joints of a robot, as if the robot is an extension of the human body. The robot follows actions of the expert to complete a task, during which the actions as well as other environmental information are recorded as demonstrations for later robot training.

**3.1 Development Platform and Setup**

In this study, a widely applied 6-degree-of-freedom robot UR3 was used. It has six revolute joints and could be equipped with multiple end effectors. UR3 has a payload of 3kg and a reach radius of 500mm. The rotation degree for each revolute joint is [-180°, 180°]. It should be noted that the modules developed in this study applies to any 6-degree-of-freedom robot arms, and UR3 was chosen because of its wide adoption.

The VR environment was established in Unity 3D [56]. Unity is a cross-platform game engine developed for 2D and 3D scenario build. The VR scenario is presented on an Oculus Quest 2 device. Oculus Quest 2 has 4 infrared cameras in the front of the headset to track the user and the movable space of the user. Since VR devices are merely containers of the virtual environment, the VR environments can also be placed in other VR devices.

**3.2 Gesture Recognition and Action Mapping**

In this module, we suggested that instead of controlling the robot via VR controllers, directly mapping the position, rotation, and actions of the hand to the robot end effector will be a more intuitive approach. The first step is to utilize the VR headset to track the position and rotation of the joints of our hands. With the skeleton data of the hand, we can define several gestures to control robot actions (e.g., open and close of the end effector), and map the position and rotation of the hand to the end effector.

Hand tracking is achieved via the four infrared cameras in the front of the headset. This module aims to recognize several specific gestures and pass corresponding commands to the UR3 robot to complete gripping, screwing, and suction tasks (Fig. 2).

*Gripping task*: The robot arm first needs to approach the target, and when the closing command is received, the end effector (i.e., parallel gripper) will close at a constant speed until it touches the object. When the opening command is received, the end effector will open its grippers at a constant speed until it reaches the maximum distance. When the fixed command is received, the end effector will hold the current pose.

*Screwing task*: First, the robot arm should grip the object using the gripper (gripping tasks), and when the tightening command is received, the end effector will spin to tighten the object. Similarly, the stopped command will cancel the screwing action, the loosening command will let the end effector spin to loosen the object, and the fixed command will let it hold its current pose.

*Suction tasks*: Similar to the gripping task, four suction cups execute the pick and place actions via suctioning and dropping command, and the fixed command just make suction cups hold their current poses.

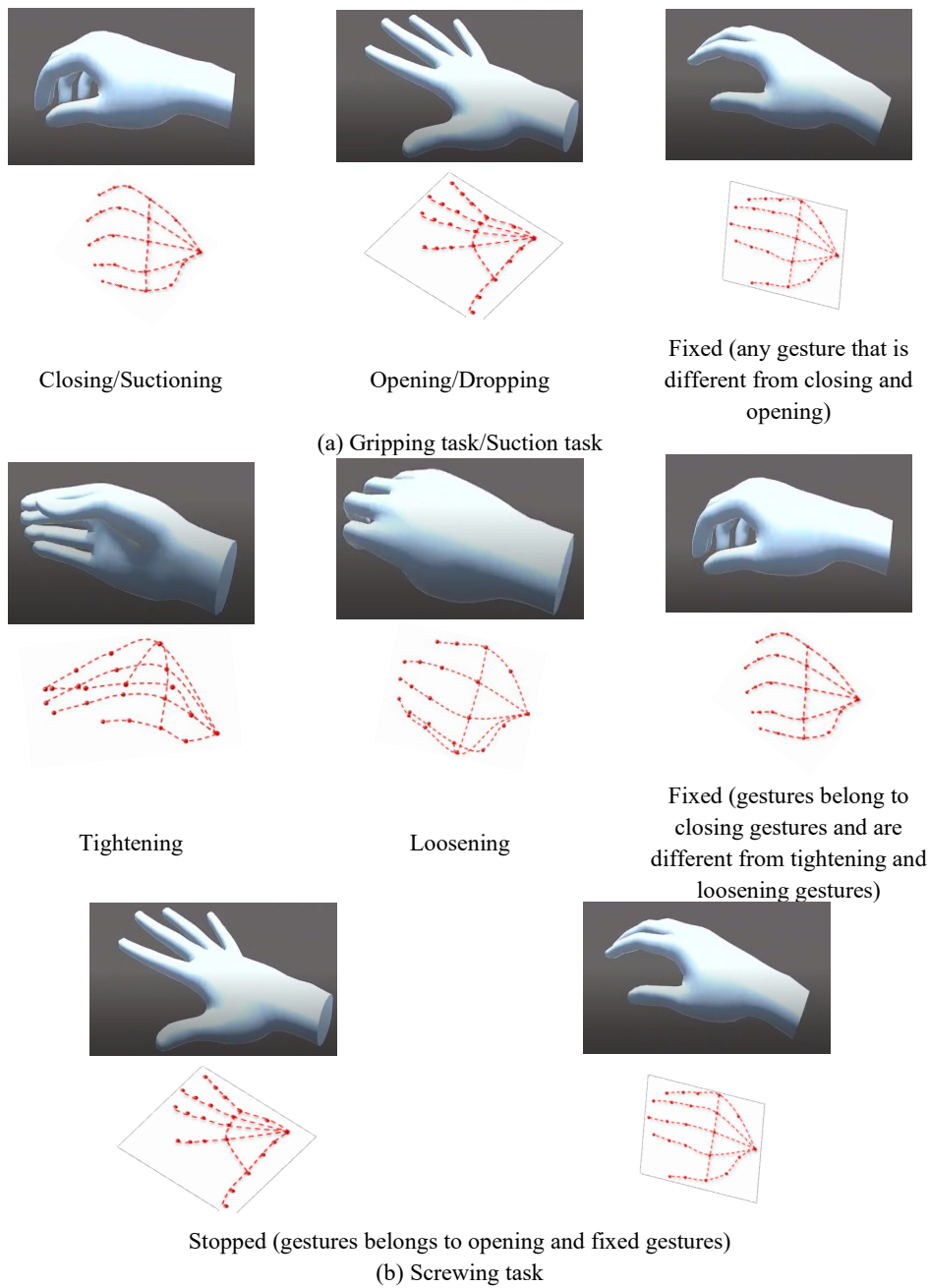

Fig. 2 Three basic construction tasks and corresponding gestures.

For the gesture recognition model, the inputs are the x, y, and z coordinates of the skeleton, and the outputs are the corresponding classes of gestures (i.e., opening, closing, and fixed). For this gesture recognition model, support vector machine (SVM) was used since SVM has been demonstrated to have remarkable performance on small and structured datasets while consuming fewer computational resources [57].

It should be noted that there is no difference between gripping and suction in terms of gesture representation since they just use different end effectors to finish similar tasks. For gripping and suction tasks, it is intuitive for demonstrators to utilize the grasp gestures (Fig. 2) to invoke the end effector.

The screwing task, however, is more complex as we use a parallel gripper to complete the task. The gestures for the gripping task are also included in the screwing task since we need to grip the object and then, complete the task by spinning the gripper, as is shown in Fig. 3. An appropriate method for this problem is that the tightening, fixed, and loosening gesture are the subset of the closing gesture and the stopped gesture involves the opening and fixed gestures of the gripping task (Fig. 3), this can be expressed by a tuple composed of two gestures (gripping gestures, screwing gestures). Thus, only after the object is gripped by the gripper, the gripper can tighten or loosen it with its current closing state.

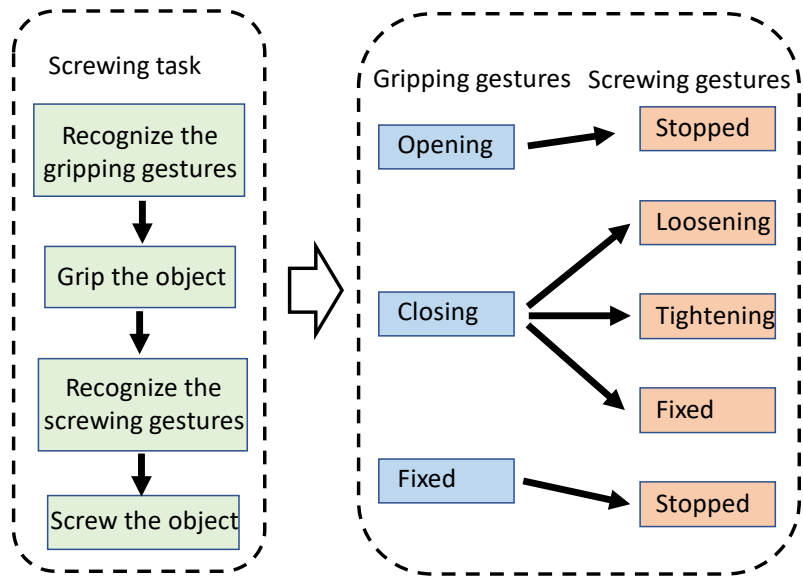

Fig. 3 The workflow of the screwing task and the definition of screwing gestures. The gestures that open the gripper or keep the gripper fixed will not screw the object (stopped gesture). Since we can screw the object only when the gripper is closed, the gestures that close the gripper are classified into three groups: loosening, tightening, and fixed (just keep the current state).

For gripping and suction tasks, 827 samples (i.e., skeleton data) were collected where 20% were utilized for testing and 80% were used for training. The collection process is shown in Fig. 4. Samples belonging to three poses were 290 (closing/suctioning), 274 (opening/dropping), and 263 (fixed), respectively. For the screwing task, 6925 samples were collected where 30% were utilized for the test and 70% were used for training. Samples belonging to five gesture states were 1475 (closing, tightening), 1289 (closing, loosing), 1494 (closing, fixed), 1209 (opening, stopped), and 1458 (fixed, stopped), respectively.

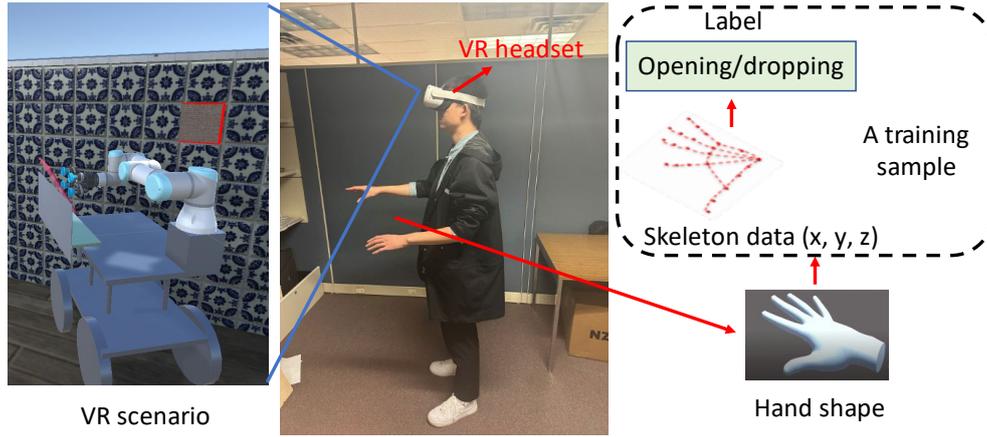

Fig. 4 The collection of samples. We get the skeleton data of the right hand and manually label the data for training.

5-fold cross-validation was applied to the training datasets for hyper-parameter tuning. The final results are shown in Table 1. Results show that SVM models have high accuracy on all tasks, so it is reliable to apply SVM models to gesture recognition.

**Table 1 Performance of SVM models.**

|  | Test dataset | Accuracy | Corresponding tasks |
| --- | --- | --- | --- |
| SVM model 1 | 165 samples | 98.19% | Gripping and suction tasks |
| SVM model 2 | 2078 samples | 99.28% | Screwing task |

Since the UR3 robot has more links (the physical connection between joints) than a human arm, it is reasonable to only map the position and rotation of the hand to the end effector, and solve the position and rotation of the UR3 joints through inverse kinematics, as shown in Fig. 5. Defining $\underline{y}'$ as a vector from the position where the wrist is located to the position where the middle proximal phalange bone is located, $\underline{x}'$ as a vector from the position where the index proximal phalange bone is located to the position where the middle proximal phalange bone is located, the $z$-axis can be represented by the cross-product of $\underline{x}'$ and $\underline{y}'$, i.e., $\underline{z} = \frac{\underline{x}' \times \underline{y}'}{\|\underline{x}' \times \underline{y}'\|}$. The $y$-axis should be $\underline{y} = \frac{\underline{y}'}{\|\underline{y}'\|}$, and then, the x axis can be represented by $\underline{y} \times \underline{z}$. Since $\underline{x}'$ is not perpendicular to $\underline{y}'$, the aforementioned method can ensure a unit orthogonal coordinate system. Therefore, assuming the coordinate ($x$, $y$, $z$) where the wrist is located is the target position of the end effector, we can compute the target transformation matrix $^0T_{6,\,target}$ of the end effector with respect to the base coordinate system with target coordinate and target coordinate system (i.e., the representation of $(\underline{x}, \underline{y}, \underline{z})$ with respect to the base coordinate system). Next, the rotation of each joint can be calculated via inverse kinematics with $^0T_{6,\,target}$.

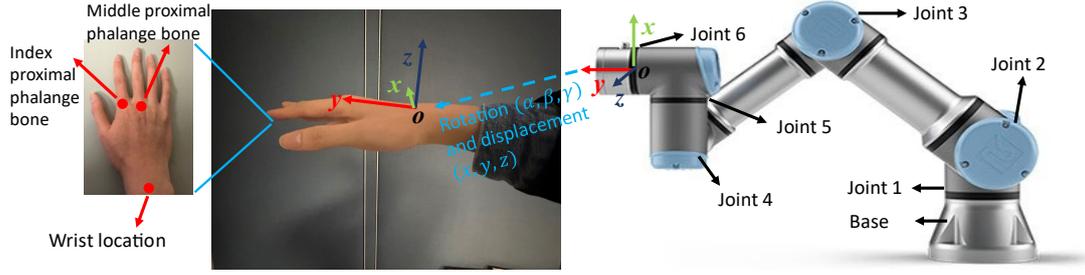

Fig. 5 Mapping from the human hand to a UR3 robot.

### 3.3 Inverse Kinematics of the UR3 robot

Given the position $(x, y, z)$ and orientation $(\alpha, \beta, \gamma)$ of the end effector coordinate system, the goal of inverse kinematics is to provide the rotation or displacement of revolute or prismatic joints. The first step is to calculate the target transformation matrix $^0T_{6,\text{target}}$ of the end effector with respect to the base coordinate system, as shown in Eq. (1).

$$^0T_{6,\text{target}} = \begin{bmatrix} \underline{x}_1 & \underline{y}_1 & \underline{z}_1 & x \\ \underline{x}_2 & \underline{y}_2 & \underline{z}_2 & y \\ \underline{x}_3 & \underline{y}_3 & \underline{z}_3 & z \\ 0 & 0 & 0 & 1 \end{bmatrix} \quad (1)$$

Where the subscript $i$ of $\underline{x}, \underline{y},$ and $\underline{z}$ denotes the $i$th element of the vector. The physical meaning of $\underline{x}, \underline{y},$ and $\underline{z}$ is the representation ($x$, $y$, and $z$ coordinates) of the end effector coordinate system with respect to the base coordinate system. It is convenient to calculate the current transformation matrix $^0T_{6,\text{current}}$ via forward kinematics, so the goal is to minimize the difference between $^0T_{6,\text{current}}$, and $^0T_{6,\text{target}}$.

Assuming $\underline{J}(q)$ denotes the Jacobian matrix where $q$ represents the rotation angle for each revolute joint, $\underline{J}(q)$ expresses the influence of the rate of the change of joint angles on that of the end effector, which can be written as:

$$\begin{bmatrix} \underline{\dot{o}}_n \\ \underline{w}_n \end{bmatrix} = [\underline{J}_0 \cdots \underline{J}_{n-1}](q)\dot{q} \quad (2)$$

Where $\underline{\dot{o}}_n$ represents the origin of the $n$th coordinate system; $\underline{w}_n$ denotes the angle velocity of the end effector; $[\underline{J}_0 \cdots \underline{J}_{n-1}](q)$ is the Jacobian matrix and can be denoted by $\underline{J}$. Eq (2) implies that $\dot{q}$ is equal to $\underline{J}^{-1}\begin{bmatrix}\underline{\dot{o}}_n \\ \underline{w}_n\end{bmatrix}$, i.e., $\nabla q = \underline{J}^{-1}\nabla\begin{bmatrix}\underline{o}_n \\ \underline{\theta}_n\end{bmatrix}$. The initial state of joints is $q_0$ and the displacement and rotation (i.e., $\nabla\begin{bmatrix}\underline{o}_n \\ \underline{\theta}_n\end{bmatrix}$) from current transformation matrix $^0T_{6,\text{current}}$ to target transformation matrix $^0T_{6,\text{target}}$ are known quantities, so we need to calculate the target state $q$ of joints. According to Eq (2), this can be solved by the Gauss-Newton method. The algorithm is shown in Table 2.

Table 2 Inverse kinematics for UR3 robot.

**Algorithm:** Inverse kinematics for UR3 robot.
**Input:** initial joint angles $q_0$, target position $p$ and target rotation $r$, the DH parameter table of the UR3 robot.
**Output:** target joint angles $q$.
Computing $^0T_{6,\text{ target}}$ with $p$ and $r$ by Eq. (1).
Computing $^0T_{6,\text{ current}}$ by forward kinematics.
Computing error $e = \|^0T_{6,\text{ target}} - {}^0T_{6,\text{ current}}\|$.
**While** $e > e_{max}$:
    Computing $J$ with $q_i, i = 0,1,2,...$
    Computing the angle-axis expression $g$ of transformation from $^0T_{6,\text{ current}}$ to $^0T_{6,\text{ target}}$.
    **if** $J$ is invertible:
        Computing $q_{i+1}$ with $g$ and $J^{-1}$ using the Gauss-Newton method.
    **else**:
        Computing $q_{i+1}$ with $g$ and the pseudo-inverse of $J$ using the Gauss-Newton method.
    Computing $^0T_{6,\text{ current}}$ by forward kinematics with $q_{i+1}$.
    Modifying the computed angles in $q_{i+1}$ to $[-180°, 180°]$.
    Computing $e$.

## 4. Imitation learning-based control module

This module aims to provide a platform where the demonstrations collected in the previous module can be utilized to train RL models. The RL model is expected to control the UR3 robot to complete construction tasks.

In terms of reward function design, rewarding states is better than rewarding actions [56]. Therefore, the robot will get a reward only if it completes subtasks, which raises a sparse reward problem. IL methods such as GAIL and BC have been manifested to have outstanding performance in sparse reward problems. In addition, given that human actions might not be the perfect trajectories for the tasks, the PPO algorithm was also added after warm-up from BC to encourage exploration. As a result, PPO, GAIL, and BC were combined to perform the control. To elaborate, the combination policy is as follows: first, BC is used for pre-training of the policy at the early stage; after that, GAIL and PPO are used for training. GAIL forces the robot control policy to imitate expert policy, which provides the intrinsic reward (rewarding the similarity between the expert policy and the agent policy), and PPO focuses on the environmental feedback (rewarding interactions with the environment), which is defined as the extrinsic reward. We used the strength coefficient to adjust the weights of different rewards along with different algorithms' influence on the policy.

### 4.1 Proximal Policy Optimization

PPO aims to maximize the performance measure function $J(\theta)$ by optimizing a policy $\pi_\theta(a|s)$ where state $s$ is the input and the robot action $a$ is the output of the neural network parameterized by $\theta$. Therefore, according to the gradient descent method, this process depends on the calculation of $\nabla J(\theta)$, which can be written as [19]:

$$\nabla J(\theta) \propto \mathbb{E}_\pi q_\pi(s,a) \nabla_\theta \ln \pi(a|s,\theta) \quad (3)$$

Where $\mu(s)$ is the stationary distribution of the Markov chain for policy $\pi$ and $q_\pi(s,a)$ denotes the value that a robot can obtain at state $s$ and by executing action $a$. To reduce variance, $q_\pi(s,a)$ can be replaced by $\hat{A}(s,a)$ [58]. $\hat{A}(s,a)$ is equal to $q_\pi(s,a)$ minus $V_\pi(s)$, where $V_\pi(s)$ is the value that a robot can obtain from state $s$. To improve the stability, the policy should not change drastically in one step. PPO uses a clip function to apply the constraint on the update. This term can be written as [58]:

$$L^{CLIP}(\theta) = \mathbb{E}[\min(\frac{\pi_\theta(a|s)}{\pi_{\theta_{old}}(a|s)}\hat{A}(s,a), clip(\frac{\pi_\theta(a|s)}{\pi_{\theta_{old}}(a|s)}, 1-\zeta, 1+\zeta)\hat{A}(s,a))] \quad (4)$$

Where $clip(\cdot)$ modifies $\frac{\pi_\theta(a|s)}{\pi_{\theta_{old}}(a|s)}$ to $[1-\zeta, 1+\zeta]$, and $\pi_{\theta_{old}}(a|s)$ denotes the policy before the update. Meanwhile, combining a value function error term $L^{VF}(\theta)$ and $L^{CLIP}(\theta)$, PPO enables the policy and value function to share parameters via a neural network architecture. To encourage exploration, PPO also adds an entropy bonus in the final loss function $L^{CLIP+VF+S}(\theta)$. $L^{CLIP+VF+S}(\theta)$ can be written as [58]:

$$L^{CLIP+VF+S}(\theta) = \mathbb{E}[L^{CLIP}(\theta) + c_1 L^{VF}(\theta) + c_2 S[\pi_\theta](s)] \quad (5)$$

Where $c_1$ and $c_2$ represent the weights of the value function error term and the entropy bonus, respectively; $L^{VF}(\theta)$ is the square error between the output of the value function and the target value; $S[\pi_\theta](s)$ is the entropy bonus, it will reach its maximum when the probability distribution of actions at each state is uniform.

Based on the above discussion, for each iteration, PPO runs the policy $\pi_{\theta_{old}}$ for $T$ timesteps for each actor and calculate the advantage estimates $\hat{A}$. Then, for each iteration, PPO will update $\theta$ by optimizing $L^{CLIP+VF+S}(\theta)$ with $K$ epochs and minibatch size $M$ [58].

**4.2 Generative Adversarial Imitation Learning**

GAIL [28] aims to optimize a policy similar to the policy extracted from expert demonstrations via IRL. Generally, IRL is concerned with parameterizing the reward function from a set of basis functions (e.g., linear combinations of basis functions $f_1, f_2, ...$), and this reward function should make the expert policy the optimal policy while maximizing the difference between the expert policy and the suboptimal policy [20]. Since only a set of trajectories sampled by executing expert policy $\pi_E$ are accessible, the idea is to compute the output of the reward function of $\pi_E$ by calculating the expected value from the trajectories [31]. Then, this is followed by a regular RL procedure to generate the expert policy by maximizing the learned reward function [32].

Instead of learning a reward function that leads to low computational efficiency, GAIL directly learns a policy from expert trajectories [28]. GAIL adopts a discriminator network $D_w: S \times A \rightarrow (0, 1)$ to distinguish the occupancy measure $\rho_\pi$ of the learner and that $\rho_{\pi_E}$ of the expert policy. The occupancy measure is defined as:

$$\rho_\pi(s, a) = \pi(a|s) \sum_{t=0}^{\infty} \gamma^t P(s_t = s|\pi) \qquad (6)$$

Where $\gamma$ is the discount factor. Eq. (6) implies that the occupancy measure is the unnormalized distribution of state-action pairs $(s, a)$ that an agent encounters with the policy $\pi$. Therefore, considering Eq. (7) as the reward function, the goal is to match the occupancy measure of agent policy $\pi$ with the expert occupancy measure (computed from demonstrations), i.e., minimize the reward function Eq. (7) using PPO rules with the $D_w$ that maximizes Eq. (8) [28].

$$c(s, a) = \log D_w(s, a) \qquad (7)$$

$$\mathbb{E}_\pi[\log D_w(s, a)] + \mathbb{E}_{\pi_E}[\log(1 - D_w(s, a))] - \lambda H(\pi) \qquad (8)$$

Where $\lambda H(\pi)$ is the regularization term and $H(\pi)$ is the causal entropy [33] of policy $\pi$. Therefore, for each iteration, GAIL samples expert trajectories $\tau_i$, and updates parameters of $D_w$ from $w_i$ to $w_{i+1}$ with Eq. (8). Then, GAIL updates the policy by using PPO in which Eq. (7) is the reward function to update $\theta_i$ to $\theta_{i+1}$.

### 4.3 Behavioral Cloning

BC aims to learn a policy from expert demonstrations via supervised learning [30]. The observations are utilized as the input of the model and the actions are the output [26]. In this module, BC is selected as the pre-training part of IL.

## 5. Experiments and Results

### 5.1 Task Setup

In this paper, we selected the suction task as an example to show how this module works. In the task, a UR3 robot equipped with four suction cups is tasked with installing tiles (Fig. 6). First, the UR3 robot needs to approach the tile. When the distance between the center of the suction cups and the tile is less than the threshold (0.04m in this case) and the robot receives the suction command, the robot can pick up the tile. Then, the robot needs to carry the tile to the target position, and if the distance between the tile and the target position is less than the threshold (0.03m in this case), the task is finished. The scenario is shown in Fig. 6.

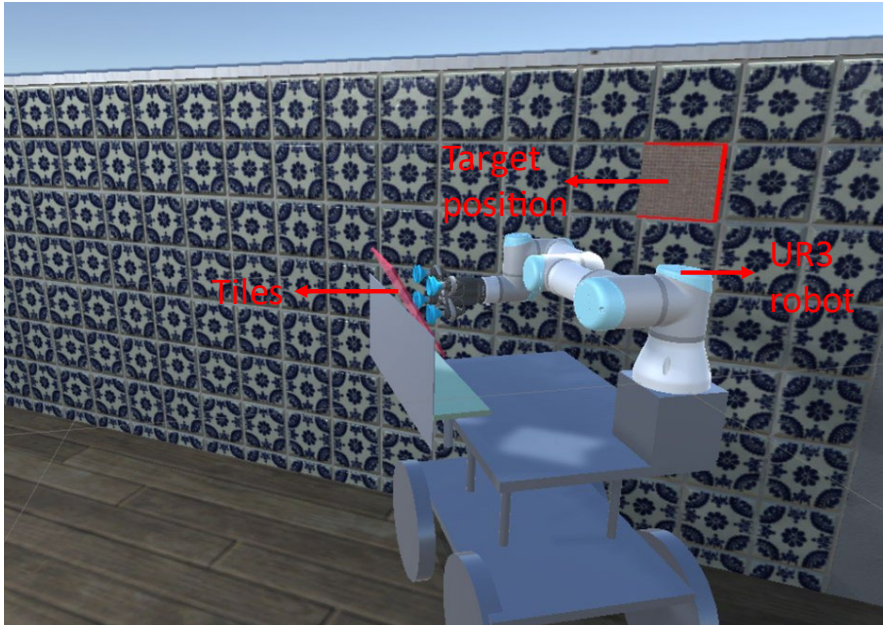

Fig. 6 The tile installation task.

The reward schedule is as follows: when the tile is picked, the agent will receive a 1.0 reward. When the robot carries the glass to the target position, the episode is completed and the agent will receive another 1.0 reward. When the tile falls, the agent will receive a -0.5 reward and the episode will be terminated. The reward function can be written as:

$$r(s,a) = \begin{cases} 1.0 & Tile\ suction \\ 1.0 & Tile\ installation \\ -0.5 & Tile\ falls \end{cases} \quad (9)$$

For each episode, the $x$ coordinate of the tile is randomized between 0 and 0.15m to ensure the agent learns a robust policy. To complete the task, the robot needs to know its joint rotation, the action of the end effector, the coordinates of the end effector, the target position, and the tile position. The action space includes the rotation of six joints (from -180° to 180°) and the actions of the end effector. Given that actions and observations are continuous sequences, we stacked actions and observations in five steps, and for the first four steps, the actions and observations are padded with zeroes.

Seven groups of experiment (Table 3) were designed to investigate the effectiveness and influence factors of IL. Groups 1, 2, and 4 were applied to study the selection of RL methods. Groups 3~7 were used to study the influence of the relative strength of PPO, GAIL, and BC. Groups 4, 8, and 9 are designed to investigate the influence of demonstration dataset size. Table 4 lists the configuration of PPO, GAIL, and BC. When the training step reaches 5,000,000 or converges to the maximum reward, the training is stopped.

Table 3 Group design for influence factor analysis.

|  | Algorithms | Demonstrations |
|---|---|---|
| Group 1 | PPO | - |
| Group 2 | GAIL | 60 |
| Group 3 | PPO (1.0*), GAIL (0.005*), BC (0.5*) | 60 |
| Group 4 | PPO (1.0*), GAIL (0.01*), BC (0.5*) | 60 |
| Group 5 | PPO (1.0*), GAIL (0.1*), BC (0.5*) | 60 |
| Group 6 | PPO (0.5*), GAIL (0.5*), BC (0.5*) | 60 |
| Group 7 | PPO (0.1*), GAIL (1.0*), BC (0.5*) | 60 |
| Group 8 | PPO (1.0*), GAIL (0.01*), BC (0.5*) | 20 |
| Group 9 | PPO (1.0*), GAIL (0.01*), BC (0.5*) | 10 |

* Means the strength of methods. For example, if the strength of PPO is 1.0, then the extrinsic (environment) reward will be multiplied by 1.0.

Table 4 The configuration of PPO, GAIL, and BC.

| PPO | GAIL | BC |
|---|---|---|
| Batch size: 512 | $\gamma$: 0.99 | Steps: 100,000 |
| Buffer size: 10,240 | Learning rate: 0.0005 |  |
| Learning rate: 0.0005 | Network settings: |  |
| $\zeta$: 0.2 | Hidden units: 256 |  |
| $c_1$: 0.95 | Number of layers: 3 |  |
| $c_2$: 0.01 |  |  |
| Horizon length: 400 |  |  |
| Network settings: |  |  |
| Hidden units: 256 |  |  |
| Number of layers: 3 |  |  |

**5.2 Results and discussions**

The learning curves for each group are shown in Fig. 7. The curve Group1 (Fig. 7a) shows the training with the PPO algorithm. It is noted that only with PPO, the robot cannot learn the skill within limited steps. The learning curve is relatively flat with the majority of the values below 0. Since there does not exist a constant punishment (it will be convenient to distinguish between different states such as tile installation, tile picked, and fall without the constant punishment), 0 reward means the robot has not touched the tile during the entire episode. This curve implies that during training, actions taken by the robot is to stay away from the tile as it finds it is much easier to knock the tile off and get a negative reward due to the low threshold (0.04m) for picking. This learning curve reveals that plain policy gradient methods such as PPO are not appropriate for solving sparse reward tasks in limited steps due to the inefficiency in searching for effective experiences.

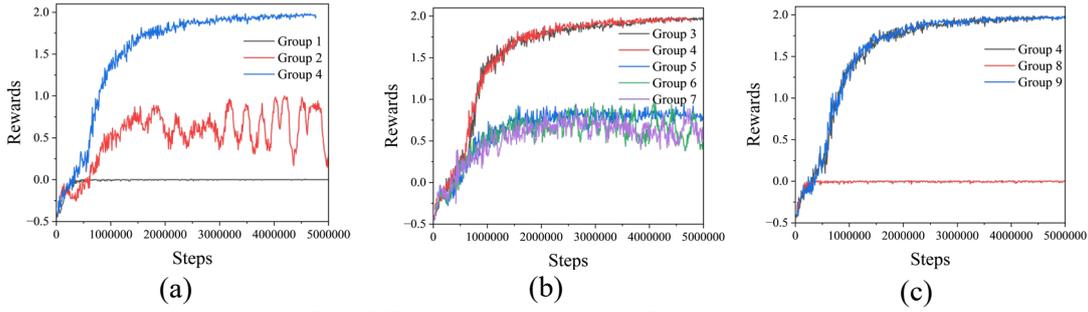

Fig. 7 The learning curve of RL models.

The curve Group 2 (Fig. 7a) shows the learning performance with the GAIL algorithm. Compared with group 1, group 2 performs better. As an IL algorithm, GAIL forces the robot to modify its policy so that the occupancy measure of its policy is similar to that of the expert policy. In this experimental test, the robot quickly obtained the skill to pick the tile. However, the curve also implies that the robot failed to carry the tile to the target position within limited steps, i.e., it either dropped the tile or held it until the end of an episode.

Groups 3~9 combine PPO, GAIL, and BC for learning. The combination of RL methods provides similar or better performance in comparison with groups 1~2 (Fig. 7a). Groups 3~7 investigate the influence of the strength coefficients. Results demonstrate that although group 1 with plain PPO shows the lowest reward, adequate exploration with slight IL constraints (high strength for PPO and low strength for GAIL) leads to better learning performance in comparison with strong IL with slight exploration rewards (low strength for PPO and high strength for GAIL), as shown in Fig. 7b. Their reward functions account for this phenomenon: a relatively high IL reward might not lead to success in the sparse reward task (e.g., the suction cup is close to the tile then it gets a high imitation reward but the distance is still not enough for picking so it cannot complete the task) but the robot can get a high environmental reward when it completes a subtask, so strong IL with slight exploration rewards reduces the difference between a sparse reward task is finished and a sparse reward task has a high probability to be completed in comparison with the adequate exploration with slight IL constraints plan.

The average GAIL loss of group 2~4 for the last 100,000 steps are 0.73, 0.02, and 1.24, respectively. The GAIL loss reflects the similarity between the agent policy and the expert policy. Both groups 3 and 4 have high extrinsic rewards but group 2 have a low extrinsic reward, as is shown in Figs. 7 a and b. Since a high extrinsic reward implies that the sparse reward task is completed and there is no distinguished gap between groups 2~4 in terms of the GAIL loss, we can infer that although a low GAIL loss leads to high similarity with expert policy (which can reduce random exploration), it cannot exactly guarantee the sparse rewards are received. A smaller threshold (e.g., 0.04m for picking the tile in this case) for completing the task needs a stronger exploration since IL loss cannot distinguish whether a task is complete or not and it needs more exploration after it has learned similar policy. However, it is noted that even though exploration is encouraged, IL is also necessary since plain PPO (Fig. 7 a) has the lowest extrinsic reward.

Groups 4, 8, and 9 investigate the influence of demonstrations (Fig. 7c). It is noted that a certain number of demonstrations is needed to ensure successful learning. But when the demonstrations are enough for the agent to learn the skill, more demonstrations do not lead to a significant improvements in the speed and quality of learning.

To test the learned policy, each policy was executed 100 times and the results are shown in Table 5. The results are consistent with their learning performance. The highest task completion rate and the lowest episode length of group 3 imply it learned the best policy. However, group 1 cannot finish any subtask, and group 2 can only pick the tile. Results indicate that demonstrations help robots perform better and enough exploration with slight IL constraints plan is suggested.

Table 5 The performance of different RL models on the tile installation task.

| Group name | Picked rate | Installed rate | Average episode length |
| --- | --- | --- | --- |
| Group 1 | 0% | 0% | 750 |
| Group 2 | 18% | 0% | 750 |
| Group 3 | 97% | 97% | 90 |
| Group 4 | 97% | 97% | 173 |
| Group 5 | 93% | 0% | 712 |
| Group 6 | 2% | 0% | 737 |
| Group 7 | 27% | 0% | 672 |
| Group 8 | 0% | 0% | 750 |
| Group 9 | 97% | 82% | 128 |

The trajectory generated by the best policy (Group 3) is shown in Fig. 8. First, the robot moves its end effector to the left (from Fig. 8a to Fig. 8b) meanwhile keeping an appropriate distance to the tile. Then, it picks the tile and moves towards the target position (from Fig. 8c to Fig. 8d). When the distance between the tile and the target position is less than 0.03m, the task is completed (Fig. 8e). Fig. 8 manifests that the robot has learned to pick up the tile and place it to the target position meanwhile avoiding knocking the tile off the table.

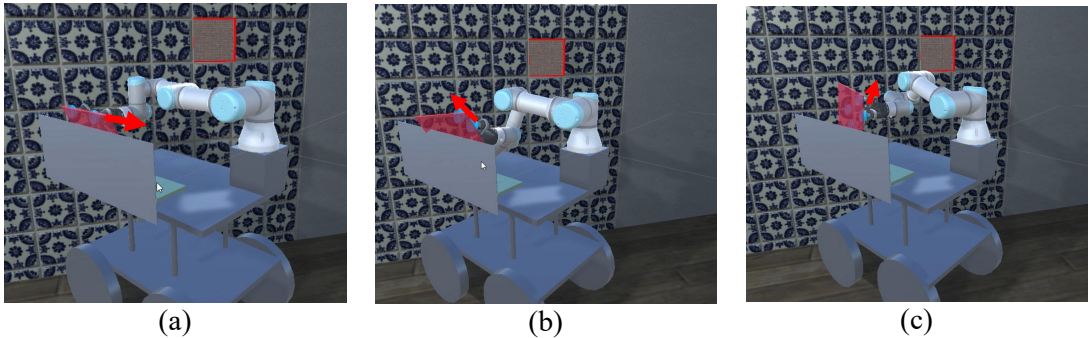

(a)　　　　　　　　　　(b)　　　　　　　　　　(c)

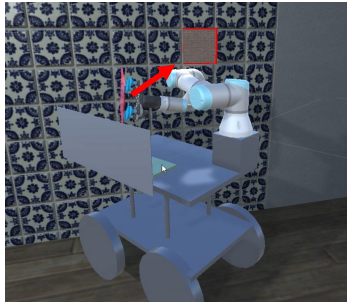 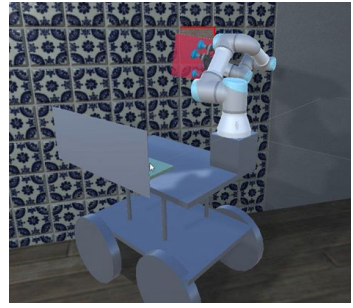

          (d)           (e)

Fig. 8 The trajectory generated by group 3.

## 6. Conclusions

Since most construction tasks involve long sequence of sub-tasks, IL has potential in improving the performance of current construction robots. As a fundamental part of IL, demonstrations are required to record the instinctive actions of experts. Current demonstration collection methods in the VR environment need the expert to control a robot via controllers, which leads to unnatural actions in demonstrations. In this paper, we provided a more intuitive alternative approach for demonstration collection. The main conclusions are as follows:

- We proposed an intuitive demonstration collection platform where VR was used to develop a realistic construction environment and the position, rotation, and actions of the hand of the expert were mapped to the end effector of a UR3 robot in real-time. This platform provides a more intuitive solution for demonstration collection as the expert does not need to concern about interactions with the VR controllers. Recording of three basic construction activities (gripping, suction, and screwing) was supported in this platform.
- Given the fact that the expert policy might not be the most efficient policy for the robot, we demonstrated that by combining GAIL, BC, and PPO, we can both accelerate the learning process within limited training steps through IL and improve the performance by allowing exploration. BC is appropriate for optimizing the policy at the early stage. A trade-off between GAIL and PPO balances the strength of imitation and exploration.
- The combination of IL and policy gradient algorithms is suggested considering the expert trajectories might not be the most efficient trajectories. The best combination plan is to allow enough exploration (high strength coefficient) while slightly modifying the policy with IL reward (low strength coefficient).

As a widely used robot, UR3 can satisfy most small payload construction tasks. The future framework will extend the current robot family to include diverse robots (e.g., UR5, UR10, UR20). In addition, given that for some actions such as screwing and gripping, the rate of the position and rotation changes of hands is time-varying and composed of a gradually varied sequence of gestures, future work will also focus on the spatial-temporal model for gesture recognition.

**Acknowledgments**

The authors acknowledge the support of the Natural Sciences and Engineering Research Council of Canada (NSERC). Kangkang acknowledges the support of the CSC scholarship from the Chinese Scholarship Council.

**Data Availability Statement**

Data generated or analyzed during the study are available from the corresponding author by request.